# Recognition of Acoustic Events Using Masked Conditional Neural Networks

Fady Medhat    David Chesmore    John Robinson
*Department of Electronic Engineering*
*University of York*
York, United Kingdon
{fady.medhat, david.chesmore, john.robinson}@york.ac.uk

*Abstract*—Automatic feature extraction using neural networks has accomplished remarkable success for images, but for sound recognition, these models are usually modified to fit the nature of the multi-dimensional temporal representation of the audio signal in spectrograms. This may not efficiently harness the time-frequency representation of the signal. The ConditionaL Neural Network (CLNN) takes into consideration the interrelation between the temporal frames, and the Masked ConditionaL Neural Network (MCLNN)[1] extends upon the CLNN by forcing a systematic sparseness over the network's weights using a binary mask. The masking allows the network to learn about frequency bands rather than bins, mimicking a filterbank used in signal transformations such as MFCC. Additionally, the Mask is designed to consider various combinations of features, which automates the feature hand-crafting process. We applied the MCLNN for the Environmental Sound Recognition problem using the Urbansound8k, YorNoise, ESC-10 and ESC-50 datasets. The MCLNN have achieved competitive performance compared to state-of-the-art Convolutional Neural Networks and hand-crafted attempts.

## I. INTRODUCTION

The sound recognition problem has been an active area of research for decades. Several feature extraction techniques and recognition models have been proposed in an attempt to tackle the problem. The feature extraction involves finding the most prominent features that can enhance the accuracy of the model used. The process of hand-crafting the features is a time-consuming stage that requires experimentation with a wide variety and combination of features to find the most effective ones for the recognition task.

Neural networks based architectures are currently being considered to automate the feature extraction phase, where deep architectures of neural networks are used to extract high-level abstract representations that can be classified by a conventional classifier e.g. SVM [1]. These architectures were adapted for different sound recognition tasks such as speech [2, 3], music [4, 5] and environmental sounds [6, 7]. Spectrograms are dominantly used as an intermediate time-frequency representation for most of the works specially to fit the 2-dimensional input expected by models based on Convolutional Neural Networks (CNN) [8] that gained a wide attention with the work of Krizhevsky et al. [9] for image classification. Dieleman et al. [10] had an attempt to eliminate the need for these intermediate representations for sound by using the raw signal, but their work shows that the frequency domain yields better performance.

[1] Code: https://github.com/fadymedhat/MCLNN
This work is funded by the European Union's Seventh Framework Programme for research, technological development and demonstration under grant agreement no. 608014 (CAPACITIE).

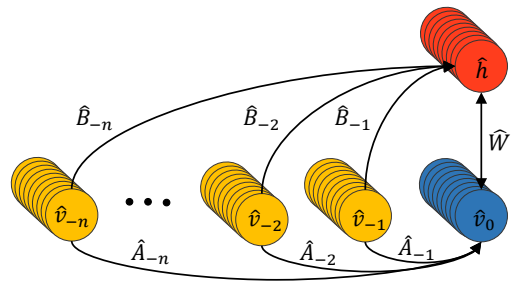

Fig. 1. The Conditional RBM structure

## II. RELATED MODELS

Deep Belief Networks (DBN) [11] were applied to the time-frequency representation for music genre classification in [12] by stacking Restricted Boltzmann Machine (RBM) [13] layers. Taylor et al. extended upon the RBM in the Conditional Restricted Boltzmann Machine (CRBM) [14] to consider the temporal nature of the signal, where they used the CRBM to model the human motion. The CRBM is an RBM with conditional relationship between the previous visible feature vectors ($\hat{v}_{-n}, \ldots, \hat{v}_{-2}, \hat{v}_{-1}$) and both the currently visible vector $\hat{v}_0$ and the hidden layer $\hat{h}$ as shown in Fig. 1. The Interpolating CRBM (ICRBM) [15] introduced by Mohamed et al. was an extension to the CRBM to include the future frames in addition to the past ones. They applied the ICRBM to the phoneme recognition task and it achieved higher accuracy compared to both the CRBM and the RBM.

The Long Short-Term Memory (LSTM) [16], used for sequence labeling, is a Recurrent Neural Network (RNN) model introduced to tackle the problem of the vanishing and exploding gradients in RNN due to the Back-Propagation Through Time. The RNN includes the influence of the previous input state over the current input. LSTM implements this behavior using an internal memory, which allows the LSTM to consider several past states of a temporal signal. Graves et al. [17] explored several LSTM architectures for phoneme recognition.

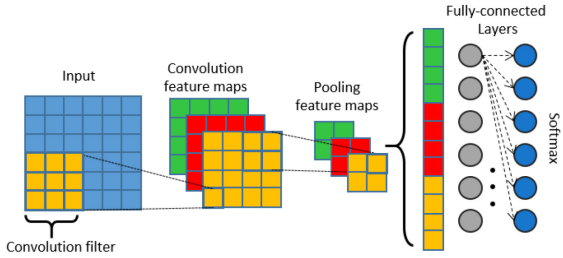

Fig. 2 Convolutional Neural Network.

Another deep model, the Convolutional Neural Network (CNN) [8], referenced earlier, uses two primary operations namely the convolution and pooling. In the convolutional operation, the input is convolved with a set of weight matrices known as filters of small sizes, e.g. 5×5. These filters scan the input image to generate new representations of feature maps, which are further pooled, either through a mean or max pooling operation, to decrease the their resolution. Several of these two operations can form a deep CNN architecture, where the feature maps generated at the last layer are flattened to form a feature vector to train a fully-connected neural network for the final classification as shown in Fig. 2.

The Convolutional RBM [18] (ConvRBM) extended the generative models family of the RBM to a Convolutional DBN (ConvDBN) by adapting the weight sharing used in the CNN and by introducing a probabilistic max-pooling layer to mimic the pooling used in the feedforward architecture of the CNN. The ConvRBM involves the use of groups of hidden layer arrays, where each group is linked to a common filter shared across the neurons of the group together with a shared bias followed by the probabilistic pooling layer. Lee et al. adopted a deep ConvDBN for audio recognition in [19].

In an attempt to exploit the performance of a hybrid model for the sound recognition task. Choi et al. [20] introduced the Convolutional Recurrent Neural Network (CRNN), where they used a CNN to extract the features and an RNN to capture the long-term dependences across the sound signal frames for music tagging.

The CNN and the models based on the convolution operation referenced earlier are designed to share the weights across different regions of a 2-dimensional input especially images. Sharing the weights does not preserve the spatial locality of the learned features, which is effective for images as it avoids the need to have a dedicated weight between the hidden layer and each pixel in the image and allows the scaling of the network for large images. This does not fit well with a spectrogram representation, where there is a need to preserve the spatial locality across both the time and frequency dimensions. Abdel-Hamid et al. [2] tackled this problem by redesigning the convolutional filters to fit the spectrogram representation of the speech signal. A similar attempt was studied in the work of Pons et al. [21], where they proposed a separate CNN architecture for each of the frequency and time domain. They achieved a higher recognition accuracy on merging the filters learned separately from each domain compared to a CNN model that is trained on the time and frequency domain concurrently. On the other hand, non-convolutional models such as DBNs treat temporal signals as separate frames, ignoring the inter-frames relation.

Different adaptations were proposed to fit the previously referenced models to the sound problem after they gain wide acceptance in domains other than sound especially images, which may not optimally harness time-frequency representations. The ConditionaL Neural Network (CLNN) [22] is designed for multi-dimensional temporal signals such as sound signals represented in a spectrogram. Extending from the CLNN, the Masked ConditionaL Neural Network (MCLNN) [22] enforces a systematic sparseness over the network weights. The sparseness enforced follows a band-like pattern, which induces the network to learn in frequency bands rather than bins, mimicking the functionality of a filterbank. The mask also plays the role of automating the hand-crafting process by considering different combinations of features concurrently. Meanwhile, the MCLNN preserves the spatial locality of the learned features. In this work, we extend the evaluation of the deep MCLNN considered in [23-25] to shallow MCLNN architectures in addition to the influence of long segments on the model's performance.

### III. CONDITIONAL NEURAL NETWORKS

The ConditionaL Neural Network (CLNN) [22] is a discriminative model stemming from the Conditional RBM. It adapts the directed links between the previous visible states and the hidden layer of a CRBM. Meanwhile, the CLNN also considers the directed links from the future frames as in the ICRBM.

For notation purposes, we will use uppercase symbols with the hat operator $\widehat{W}$ to represent matrices and lower case symbols $\hat{x}$ for vectors. A subscript index in combination with the hat operator for an uppercase symbol $\widehat{W}_u$ refers to a matrix at index $u$ in a tensor. In the absence of the hat operator we are referring to individual elements i.e. $W_{i,j}$ is the element at location $[i, j]$ of a matrix $\widehat{W}$, similarly $x_i$ is the $i^{th}$ element of the vector $\hat{x}$. Vector-matrix multiplication is referred to with the dot operator ( $\cdot$ ) and element-wise multiplication between two vectors or two matrices of the same size uses ( $\circ$ ). Absence of any operators or the use of ( $\times$ ) refers to normal element multiplication i.e. ( $x_i W_{i,j}$ or $l \times e$ ).

A single CLNN layer is formed of a vector-shaped hidden layer of $e$ neurons. The input of the CLNN is a sequence of frames forming a window of size [$l, d$], where $l$ is the number of features in a single feature vector (frame in the case of a spectrogram), and $d$ is the number of frames in the window. The width $d$ of a window follows (1)

$$d = 2n + 1 \quad , n \geq 1 \qquad (1)$$

where $d$ is the width of the window and consequently the number of frames in the window. The order $n$, a tunable hyper-parameter, specifies the number of frames that are considered in either direction of the window's central frame (2 is for the past and future frames and 1 is for the window's central frame). For each frame in the window, there is a set of full connections between each feature in a single input frame and each node in the hidden layer as depicted in Fig. 3. The figure shows the $2n+1$

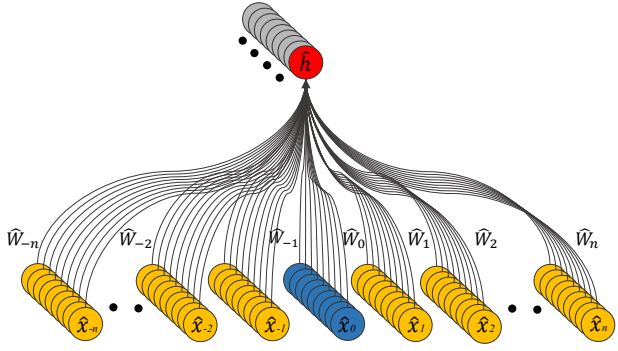

Fig. 3. ConditionaL Neural Network layer. Connections depicted are for a feature vectors $[x_{-n}, \ldots, x_{-2}, x_{-1}, x_0, x_1, x_{-2}, \ldots, x_n]$ belonging to the window, where each feature vector is fully connected to the hidden layer through its corresponding weight matrix $\widehat{W}_u$. The index $u$ is the position of the matrix in the weight tensor ranging within the interval $[-n, n]$. The output of a single neuron of the hidden layer is formulated in (2)

$$y_{j,t} = f\left(b_j + \sum_{u=-n}^{n}\sum_{i=1}^{l} x_{i,u+t}\, W_{i,j,u}\right) \quad (2)$$

where $y_{j,t}$ is the output of neuron $j$ of the hidden layer, $f$ is te transfer function applied to the input of the neuron, $b_j$ is the bias at the node, $x_{i,u+t}$ is the $i^{th}$ feature of the input feature vector $x$ at index $u$ of the window of frames (the $u$ ranges between $[-n, n]$, which are the indices of the frames in the window of width $d$ following (1)) and $W_{i,j,u}$ is the weight between the $i^{th}$ feature of the input feature vector of length $l$ at index $u$ of the window and the $j^{th}$ hidden node $y_j$. The index $t$ in $x_{i,u+t}$ and $y_{j,t}$ refers to the window's central frame. The central frame together with its neighbouring $2n$ frames ($n$ past and $n$ future frames) are processed from a chunk of the spectrogram. We will refer to this chunk as the segment (discussed later in detail), which is larger than the window in width i.e. has more number of frames. Each frame of the segment at index $t$ is the window's middle frame. The output vector of a CLNN is formulated in (3)

$$\hat{y}_t = f\left(\hat{b} + \sum_{u=-n}^{n} \hat{x}_{u+t} \cdot \widehat{W}_u\right) \quad (3)$$

where the activation vector $\hat{y}_t$ at the hidden layer of a CLNN is conditioned on the window's central frame $\hat{x}_t$ and its neighbouring frames $[\hat{x}_{-n+t}, \ldots \hat{x}_{-1+t}]$ and $[\hat{x}_{1+t}, \ldots \hat{x}_{n+t}]$ in the window. The activation vector is given by the output of the transfer function $f$. The bias vector of the hidden layer is $\hat{b}$ and $\hat{x}_{u+t}$ is the vector at index $u$ in a window of frames whose central frame is at index $t$ of the segment. $\widehat{W}_u$ is the weight matrix at index $u$, where the number of weight matrices is equal to $d$ in (1). Each $\widehat{W}_u$ has a size of [length of feature vector $l$, hidden layer width $e$]. At each index $u$ a vector-matrix multiplication is applied between each vector $\hat{x}_{u+t}$ of length $l$ and its corresponding weight matrix $\widehat{W}_u$ of size $[l, e]$. The resulting $d$ vectors, of $e$-dimensions each (following the hidden layer width), are summed dimension-wise to generate a vector to be fed to the transfer function $f$ for the hidden layer output together with the bias $\hat{b}$. The conditional distribution of the hidden activation conditioned on the window's middle frame and the $n$ frames on either of its side can be formulated in $p(\hat{y}_t | \hat{x}_{-n+t}, \ldots, \hat{x}_{-1+t}, \hat{x}_t, \hat{x}_{1+t}, \ldots, \hat{x}_{n+t}) = \sigma(\ldots)$ where $\sigma$ is a logistic transfer function (e.g. sigmoid).

It can be inferred from (3) that the output of a single CLNN layer has $2n$ fewer frames than its input. This is due to the summation applied on each feature vector at time $t$ and the $n$ frames on either of its sides. Accordingly, segments of size $[l, q]$ of the spectrogram are extracted to consider the consumption of frames through a deep architecture of a CLNN. The segment length $l$ follows the same length of a feature vector and $q$ is the width of the segment following (4)

$$q = (2n)m + k \quad , n, m \text{ and } k \geq 1 \quad (4)$$

where the number of frames in a segment of width $q$ is based on the order $n$ of the layer (2 is to account for both future and past frames), $m$ is the number of layers and the extra frames $k$ are the frames to remain after the processing applied by the CLNN. These extra frames beyond the CLNN layers can be flattened as in the feature maps of a CNN or pooled across as in [26], but here the pooling is applied feature-wise through the temporal dimension. This pooling operation behaves as an aggregation over a texture window, which was studied in [27] for music classification. Depending on the step of moving through time across a spectrogram, the overlapping number frames between a segment and another could reside in the interval $[0, q-1]$, where $q$ is the segment width (number of frames in a segment).

Several CLNN layers can be stacked to form a deep architecture, where the output of one layer introduces another form of an intermediate representation to the layer above it as shown in Fig. 4. The figure shows a model of two CLNN layers ($m = 2$), where each layer has an order $n=1$. Accordingly, each middle frame is considered with one future and one past frame. The matrix $\widehat{W}_0^b$ is the central weight matrix to process the central frame in the window of width $d$ and each of $\widehat{W}_{-1}^b$ and $\widehat{W}_1^b$ are the ones to process the off-center past and future frames, respectively, where $b$ is the layer index i.e. $b = 1, 2, …, m$. The figure also depicts the $k$ extra frames to be flattened or pooled across, where they are fed to a fully-connected neural network for the final classification decision using a softmax.

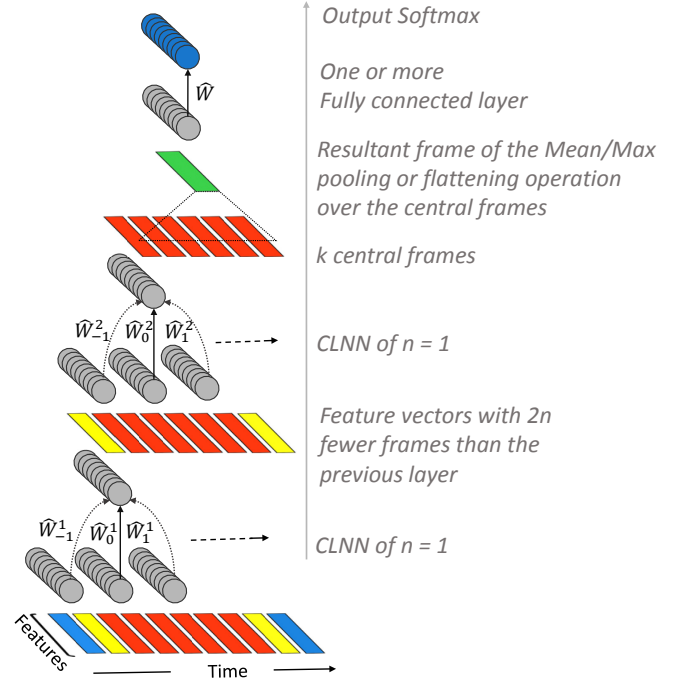

Fig. 4. Two-layer CLNN model with $n=1$

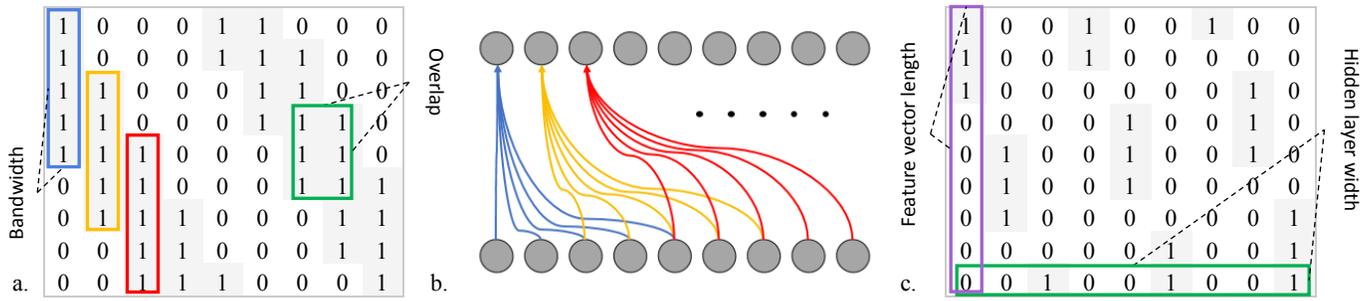

Fig. 5. Examples of the Mask patterns. a) A bandwidth of 5 with an overlap of 3, b) The allowed connections matching the mask in a. across the neurons of two layers, c) A bandwidth of 3 and an overlap of -1

## IV. MASKED CONDITIONAL NEURAL NETWORKS

The Masked ConditionaL Neural Network (MCLNN) [22] extends from the CLNN skeleton by enforcing a systematic sparseness over the network's connections aiming to embed a filterbank-like behavior within the network.

A filterbank is a group of filters used to subdivide the frequency bins of a spectrogram into bands. This transformation tackles the frequency shifts that occur in raw spectrograms due to the smearing of the energy of a frequency bin across nearby frequency bins. A filterbank allows a frequency shift-invariant representation, which provides a better comprehensible presentation of the energy across the bands as it progresses through time. The spacing between the center frequencies of the filters in a filterbank follows a specified scaling. For example, the Mel-spaced filterbank follows the Mel-scale, which is used in MFCC and Mel-scaled spectrograms.

The band-like pattern enforced over the network's weights is applied using a binary mask. Fig. 5.a. depicts an example of a binary mask, where each column has a collection of ones forming a band-like shape. The 1's positions are shifted across the columns of the mask as the figure shows. This pattern is generated by the two tunable hyperparameters that are used for the mask design, namely the Bandwidth $bw$ and the Overlap $ov$. The Bandwidth controls the count of consecutive ones across a single column, and the Overlap controls the superposition between successive columns. Fig. 5.b. shows the active connections following the Mask pattern in Fig. 5.a. The positions of the 1's are defined through a linear spacing following (5)

$$lx = a + (g - 1)(l + (bw - ov)) \quad (5)$$

where the linear index $lx$ for the position of a binary value 1 is given by the length of the feature vector $l$, the bandwidth $bw$ and the overlap $ov$. The values of $a$ range within the interval $[0, bw-1]$ and the values of $g$ are in the interval $[1, \lceil (l \times e)/(l + (bw - ov)) \rceil ]$.

The mask overlap can be assigned positive or negative values, where the negative values specify the non-overlapping distance between the successive columns as shown in Fig. 5.c. This pattern clarifies another important role for the mask, which is automating the feature combination selection process by providing several shifted versions of the filterbank-like pattern. This allows different hidden nodes to learn about different feature combinations. For example, in Fig. 5.c. the first column maps to the first hidden node. Accordingly, the 1st hidden node will learn about the first three features permitted by the 3 consecutive ones in the first column. Similarly, the 4th hidden node that maps to the fourth column will learn about the first two features of the input vector and the same applies to the 7th column allowing the node to learn about one feature only. These patterns embed a mix-and-match behavior inside the network itself, which automate the consideration of different feature combinations concurrently while preserving the spatial locality of the learned features. The masking operation is applied through an element-wise multiplication between the mask and the matrix at index $u$ belonging to the weight tensor as in (6)

$$\widehat{Z}_u = \widehat{W}_u \circ \widehat{M} \quad (6)$$

where $\widehat{W}_u$ is the original weight matrix, $\widehat{M}$ is the masking pattern and $\widehat{Z}_u$ is the new masked weight matrix to substitute the weight matrix in (3).

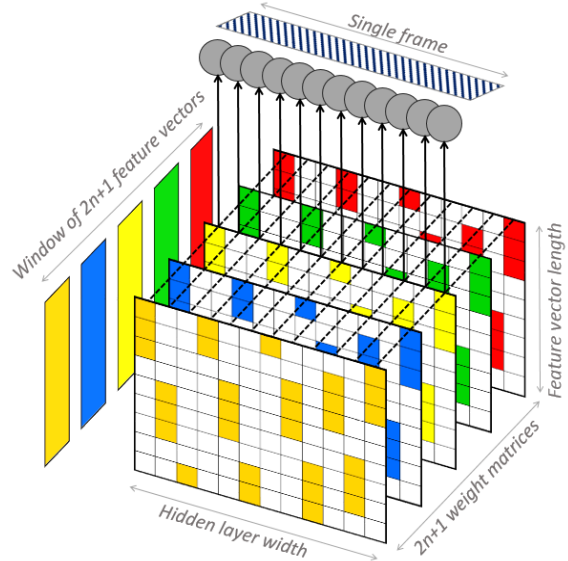

Fig. 6. A single step of MCLNN

TABLE I  MCLNN HYPER-PARAMETERS

| Layer | Type | Nodes | Mask Bandwidth | Mask Overlap | Order $n$ |
|---|---|---|---|---|---|
| 1 | MCLNN | 300 | 20 | -5 | 15 |

Fig. 6 shows a single step of the MCLNN, where $2n+1$ weight matrices are processing the $2n+1$ frames. The highlighted regions in each matrix represent the active connections following the mask design. The output of a single processing window of frames is a single representative vector.

## V. EXPERIMENTS

We have performed the MCLNN evaluation using the Urbansound8k [28], YorNoise [24], ESC-10 [29] and ESC-50 [29] environmental sound datasets. We will discuss the composition of each dataset with the common preprocessing applied, and we will defer the discussion to each dataset's relevant section. In this work, we explore the performance of a shallow architecture of the MCLNN in combination with a long segment compared to the deep MCLNN architectures considered for the mentioned datasets in [23-25].

*Urbansound8k* is composed of 8732 files for 10 classes of environmental sounds released into 10-folds: air conditioner, car horns, children playing, dog bark, drilling, idling engines, gunshot, jackhammers, siren and street music. The maximum duration for the files is 4 seconds.

*YorNoise* is a dataset focusing on rail and road traffic with 1527 sound files of 4 seconds each. The dataset is released into 10-folds following the same settings of the Urbansound8k dataset.

*ESC-10* is a dataset of 400 files of 5 seconds each for 10 categories of environmental sounds released in 5-folds: dog bark, rain, sea waves, baby cry, clock tick, person sneeze, helicopter, chainsaw, rooster and fire cracking.

*ESC-50* is a dataset of 2000 files of 5 seconds each for 50 categories of environmental sounds released in 5-folds. A subset of the classes in this dataset was used for the *ESC-10* dataset.

Common pre-processing for the datasets involved a time-frequency transformation to 60 bins logarithmically Mel-scaled spectrogram at an FFT window of 1024 and 50% overlap with the delta (first derivative across the time domain). We concatenated the spectrogram and the delta, resulting in a spectrogram frame of 120 frequency bins. The training set was z-scored, and its standardization parameters (mean and standard deviation) were applied to the validation and testing sets. The model was trained to minimize the categorical cross-entropy using ADAM [32]. We used the Parametric Rectifier Linear Units (PRelu) [33] as the transfer function for all neurons and Dropout [34] for regularization. The final decision for the sound file category is decided based on a probability voting across the predicted labels for the segments extracted from each audio file following (4). The MCLNN layer is followed by a pooling layer and two densely-connected layers of 100 neurons each before the final output softmax. Table I lists the hyperparameters used for the MCLNN. An order $n = 15$ and extra frames $k=50$ were utilized for all datasets except for the ESC-50 dataset, an order $n=14$ and $k=40$ were used.

TABLE II  PERFORMANCE ON URBANSOUND8K DATASET USING THE MCLNN COMPARED WITH OTHER ATTEMPTS IN THE LITERATURE

| Classifier and Features | Acc. % |
|---|---|
| **MCLNN (Shallow, $k$=50) + Mel-Spectrogram (This Work)** | **74.22** |
| Random Forest + Spherical K-Means + PCA + Mel-Spec.[30] | 73.70 |
| MCLNN (Deep, $k$=5) + Mel-Spectrogram [24] | 73.30 |
| Piczak-CNN + Mel-Spectrogram [7] | 73.10 |
| S&B-CNN + Mel-Spectrogram [31] | 73.00 |
| RBF-SVM + MFCC [28] | 68.00 |

|   | AC | CH | CP | DB | Dr | EI | GS | Ja | Si | SM |
|---|---|---|---|---|---|---|---|---|---|---|
| AC | **478** | 8 | 37 | 31 | 166 | 161 | 4 | 71 | 5 | 39 |
| CH | 7 | **336** | 11 | 6 | 22 | 10 | 0 | 8 | 0 | 29 |
| CP | 13 | 4 | **810** | 43 | 25 | 12 | 3 | 2 | 13 | 75 |
| DB | 19 | 12 | 72 | **837** | 8 | 5 | 6 | 3 | 13 | 25 |
| Dr | 22 | 1 | 15 | 21 | **787** | 31 | 5 | 75 | 22 | 21 |
| EI | 104 | 7 | 22 | 12 | 55 | **638** | 4 | 141 | 3 | 14 |
| GS | 2 | 1 | 0 | 17 | 2 | 2 | **349** | 0 | 1 | 0 |
| Ja | 99 | 0 | 6 | 0 | 99 | 51 | 0 | **672** | 58 | 15 |
| Si | 21 | 1 | 37 | 33 | 14 | 13 | 0 | 2 | **776** | 32 |
| SM | 23 | 3 | 126 | 22 | 7 | 1 | 0 | 7 | 13 | **798** |

Fig. 7. Urbansound8k confusion using MCLNN. Classes: Air Conditioner(AC), Car Horns(CH), Children Playing(CP), Dog Bark(DB), Drilling(Dr), Engine Idling(EI), Gun Shot(GS), Jackhammers(Ja), Siren(Si) and Street Music(SM)

### A. Urbansound8K

Environmental sound recognition research is hindered with the unavailability of a large labeled dataset. The Urbansound8K dataset was released in the work of Salamon et al. [28] in an attempt to provide a large labeled dataset for the research community. We used the model specified in Table I and the signal representation (60 mel-spec with delta) discussed earlier, which is the same transformation used by Piczak-CNN [7]. The dataset is pre-distributed into 10-folds, which we used to report the mean accuracy in Table II.

The shallow MCLNN in combination with a long segment ($k$=50) achieved an accuracy of 74.22% compared to a deep MCLNN with a shorter segment ($k$=5) in [24]. The accuracy of the MCLNN surpasses other reported neural networks based attempts using state-of-the-art CNN architectures proposed by Salamon et al. in [31] and Piczak in [7]. The baseline accuracy of 68% was achieved in [28] using an RBF-SVM [1] for classification. Salamon *et al.* in [30] achieved the highest non-neural attempt on the Urbansound8k. They proposed the use of an unsupervised feature learning technique using the Spherical k-means to establish a codebook, Principal Component Analysis (PCA) for dimensionality reduction and they used Random Forest [35] for classification. The Piczak-CNN applied through the work of Piczak et al. [7] used two convolutional layers, two

TABLE III  PERFORMANCE ON THE URBANSDOUND8K AND YORNOISE DATASETS USING THE MCLNN

| Classifier and Features | Acc. % |
|---|---|
| **MCLNN (Shallow, *k*=50) + Mel-Scaled Spectrogram** | **75.82** |
| MCLNN (Deep, *k*=5) + Mel-Scaled Spectrogram [24] | 75.13 |

TABLE IV  PERFORMANCE ON ESC-10 DATASET USING THE MCLNN COMPARED WITH OTHER ATTEMPTS IN THE LITERATURE

| Classifier and Features | Acc. % |
|---|---|
| MCLNN (Deep, *k*=40) + Mel-Scaled Spectrogram [23][2] | 85.50 |
| **MCLNN (Shallow, *k*=50) + Mel-Scaled Spectrogram (this work)[2]** | **83.00** |
| MCLNN (Deep, *k*=1) + Mel-Scaled Spectrogram [25][2] | 83.00 |
| MCLNN (Deep, *k*=25) + Mel-Scaled Spectrogram [23][2] | 82.00 |
| Piczak-CNN + Mel-Scaled Spectrogram [7][1] | 80.00 |
| Random Forest + MFCC [29][2] | 72.70 |

[1] *Augmentation*
[2] *Without Augmentation*

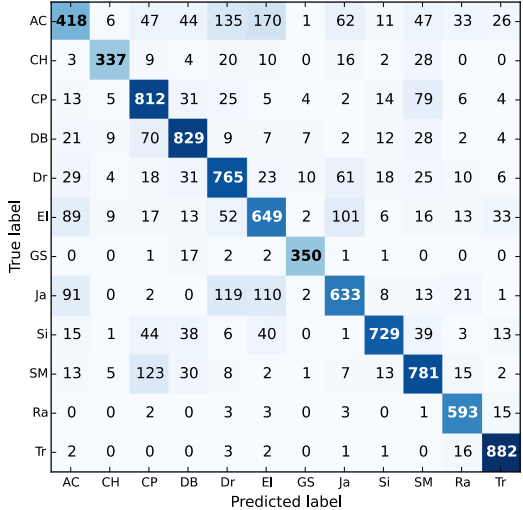

Fig. 8. YorNoise and Urbansound8k confusion using MCLNN. Classes: Air Conditioner(AC), Car Horns(CH), Children Playing(CP), Dog Bark(DB), Drilling(Dr), Engine Idling(EI), Gun Shot(GS), Jackhammers(Ja), Siren(Si), Street Music(SM), Rail (Ra) and Traffic (Tr)

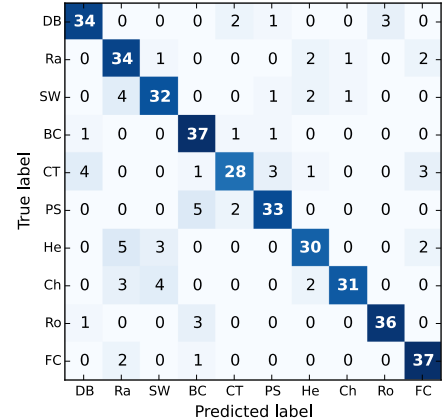

Fig. 9. Confusion matrix for the ESC-10 dataset. Classes: Dog Bark(DB), Rain(Ra), Sea Waves(SW), Baby Cry(BC), Clock Tick(CT), Person Sneeze(PS), Helicopter(He), Chainsaw(Ch), Rooster(Ro) and Fire Cracking(FC)

pooling layers and two fully-connected layers of 5000 neurons each resulting in a total number of weights exceeding 25 million. Salamon et al. in [31] used a deeper architecture then the Piczak-CNN with fewer parameters. The MCLNN achieved an accuracy of 74.22% using approximately 1 million parameters, which are less than 5% of the parameters employed in the Piczak-CNN. Fig. 7 shows the confusion across the different classes using the MCLNN. The highest confusion is occurring across the Air Conditioner, Drilling, Engine Idling and Jackhammers sounds. This is due to the high similarity of the tonal components between these categories. Similar findings were reported in the work of Salamon et al. [31] and Piczak [7].

*B. YorNoise*

The dataset is used as an extension to the Urbansound8k dataset with more emphasis on urban sounds especially rail and road traffic. The dataset is used to analyze the effect of common low tonal components across sounds generated from machines and engines on the confusion rates. The dataset has an unbalanced distribution of sound files with 620 samples for rail and 907 samples for road traffic. The YorNoise dataset is pre-distributed into 10-folds, and in combination with the Urbansound8k, it establishes a dataset composed of 12 categories of urban sounds.

Table III lists the mean accuracies achieved over a 10-fold cross-validation for both the Urbansound8k and YorNoise combined. A shallow MCLNN achieved an accuracy of 75.92% compared to the deep architecture in [24] that reached 75.13%. Despite the comparable accuracy, the shallow MCLNN used 1 million parameters compared to the 3 million parameters of the deep variant and achieved higher accuracy using a longer segment. Fig. 8 shows the confusion across the 12 classes of both datasets. The confusion extends from the machine-generated sounds of the Urbansound8k, e.g. Air Conditioner, Jack Hammer, Drilling and Engine Idling to the YorNoise Rail and Road traffic sounds due to the common tonal properties across these categories.

*C. ESC-10*

For the ESC-10 dataset, we followed the transformation applied by Piczak in [7] (60 bin Mel-spec. with Delta) to benchmark the MCLNN without the influence of the intermediate representation. The experiments followed the 5-fold cross-validation of the original distribution of the dataset to unify the reported accuracies.

Table IV lists the accuracies achieved over the dataset. The deep MCLNN architecture in [23] achieved 85.5% without augmentation with extra frames *k* = 40 and it achieved 83% at *k*=1. The shallow architecture used in this work achieved 83% using a longer segment with 1 million weights compared to the deep MCLNN that used 3 million parameters. Piczak-CNN [7] achieved an accuracy of 80% using a CNN model that used 25 million parameters (discussed in the previous section) compared to the 1 million parameters used by the shallow MCLNN. Additionally, Piczak used augmentation, which involves introducing deformations to sound signal, e.g. time delay, pitch shifting. Piczak applied 10 augmentation variants to each sound file, which increases the dataset and consequently the accuracy as studied by Salamon in [31]. We did not apply augmentation as it is not relevant to benchmarking the models we are proposing in this work. Fig. 9 shows the confusion across the ESC-10 classes using the MCLNN. The highest confusion is for the clock ticks with other short event sounds such as the person sneeze and the fire cracking sounds. There is also a noticeable confusion among the rain, sea wave, helicopter and chainsaw sounds due to the common low tones across them.

TABLE V   Performance on ESC-50 dataset using the MCLNN compared with other attempts in the literature

| Classifier and Features | Acc. % |
|---|---|
| Piczak-CNN + Mel-Spectrogram [7][1] | 64.50 |
| **MCLNN (Shallow, *k*=40) + Mel-Spectrogram (This Work)[2]** | **62.85** |
| MCLNN (Deep, *k* = 6)+ Mel-Spectrogram (This Work) [25][2] | 61.75 |
| Random Forest + MFCC [29][2] | 44.00 |

[1] *Augmentation*
[2] *Without Augmentation*

*D. ESC-50*

The dataset is pre-distributed into 5-folds. We used the same model and signal representation (60 mel-spec with Delta) we applied for the ESC-10 dataset, except for the order *n* and the extra frames *k*, where we used *n* =14 and *k*=40. Table V lists the accuracies achieved on the ESC-50 including the MCLNN. The accuracy by Piczak-CNN is based on a CNN model, described earlier, of 25 million parameters like the one applied to the ESC-10 and Urbansound8k datasets. Additionally, Piczak [7] used 4 augmentation variants for each sound file in the ESC-50 dataset. Without applying any augmentation, the MCLNN achieved 62.85% using 5% of the parameters utilized by Piczak-CNN. Fig. 10 shows the confusion across the 50 sound categories of the ESC-50 dataset.

VI. CONCLUSION AND FUTURE WORK

The ConditionaL Neural Network (CLNN) and its extension the Masked Conditional Neural Network (MCLNN) are designed for multi-dimensional temporal signals. The CLNN considers the inter-frame relation across a temporal signal, and the MCLNN extends the CLNN by enforcing a systematic sparseness through a binary mask following a band-like pattern. The mask allows the network to learn in bands rather than bins, mimicking the behavior of the filterbank used in spectrogram transformations such as Mel-Scaled analysis. Additionally, the mask is designed to include several shifted versions of the filterbank-like pattern, which automates the hand-crafting process of the feature combinations. This allows each node in the hidden layer to learn distinct localized features in its scope of observation. We benchmarked the MCLNN using the Urbansound8k, YorNoise, ESC-10 and ESC-50 environmental sounds datasets. MCLNN have achieved competitive results compared to models based on state-of-the-art Convolutional Neural Networks (CNN) in addition to hand-crafted attempts. We applied the MCLNN on a time-frequency representation, but MCLNN still preserves the generalization of applying it to other multi-dimensional representations of temporal signals, which we will explore in our future work.

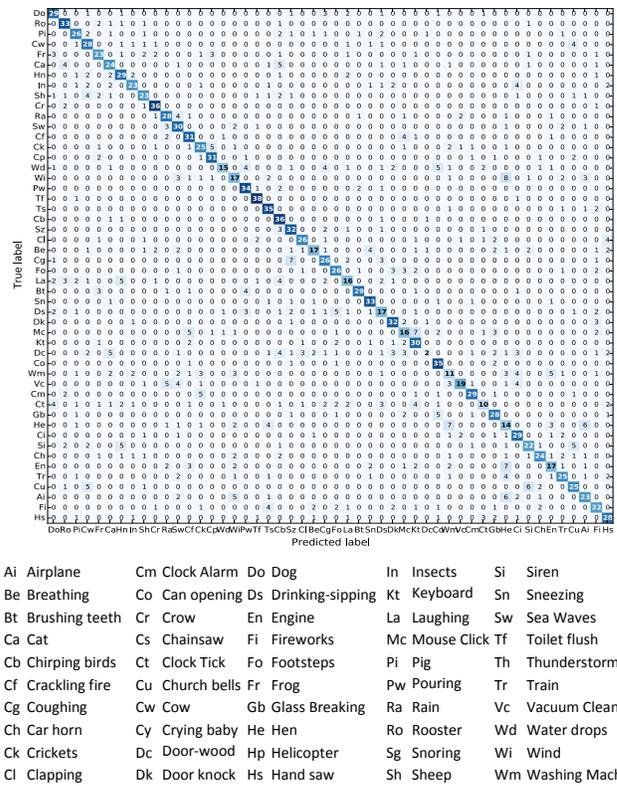

| Ai | Airplane | Cm | Clock Alarm | Do | Dog | In | Insects | Si | Siren |
| Be | Breathing | Co | Can opening | Ds | Drinking-sipping | Kt | Keyboard | Sn | Sneezing |
| Bt | Brushing teeth | Cr | Crow | En | Engine | La | Laughing | Sw | Sea Waves |
| Ca | Cat | Cs | Chainsaw | Fi | Fireworks | Mc | Mouse Click | Tf | Toilet flush |
| Cb | Chirping birds | Ct | Clock Tick | Fo | Footsteps | Pi | Pig | Th | Thunderstorm |
| Cf | Crackling fire | Cu | Church bells | Fr | Frog | Pw | Pouring | Tr | Train |
| Cg | Coughing | Cw | Cow | Gb | Glass Breaking | Ra | Rain | Vc | Vacuum Cleaner |
| Ch | Car horn | Cy | Crying baby | He | Hen | Ro | Rooster | Wd | Water drops |
| Ck | Crickets | Dc | Door-wood | Hp | Helicopter | Sg | Snoring | Wi | Wind |
| Cl | Clapping | Dk | Door knock | Hs | Hand saw | Sh | Sheep | Wm | Washing Machine |

Fig. 10. Confusion matrix for the ESC-50 using the MCLNN